\documentclass[11pt,letterpaper]{article}
\usepackage{emnlp2017}
\usepackage{times}
\usepackage{latexsym}
\usepackage{amsmath}
\usepackage{graphicx}
\usepackage{bm}
\usepackage{amssymb}
\usepackage{footnote}
\usepackage{kotex}

\emnlpfinalcopy

\def\emnlppaperid{***}

\title{Syllable-level Neural Language Model for Agglutinative Language}

\author{ Seunghak Yu\Thanks{\enskip Equal contribution} \qquad  Nilesh Kulkarni\footnotemark[1] \qquad Haejun Lee \qquad Jihie Kim \\
	Samsung Electronics Co. Ltd., South Korea\\
      {\tt \{seunghak.yu, n93.kulkarni, haejun82.lee, jihie.kim\}@samsung.com}}
\date{}

\begin{document}

\maketitle

\begin{abstract}
Language models for agglutinative languages have always been hindered in past due to myriad of agglutinations possible to any given word through various affixes. We propose a method to diminish the problem of out-of-vocabulary words by introducing an embedding derived from syllables and morphemes which leverages the agglutinative property. Our model outperforms character-level embedding in perplexity by 16.87 with 9.50M parameters. Proposed method achieves state of the art performance over existing input prediction methods in terms of Key Stroke Saving and has been commercialized.
\end{abstract}

\section{Introduction}

Recurrent neural networks (RNNs) exhibit dynamic temporal behavior which makes them ideal architectures to model sequential data. In recent times, RNNs have shown state of the art performance on tasks of language modeling (RNN-LM), beating the statistical modeling techniques by a huge margin \cite{mikolov2010recurrent,  lin2015hierarchical, kim2016character, miyamoto2016gated}. RNN-LMs model the probability distribution over the words in vocabulary conditioned on a given input context. The sizes of these networks are primarily dependent on their vocabulary size. 


Since agglutinative languages, such as Korean, Japanese, and Turkish, have a huge number of words in the vocabulary, it is considerably hard to train word-level RNN-LM. Korean is agglutinative in its morphology; words mainly contain different morphemes to determine the meaning of the word hence increasing the vocabulary size for language model training. A given word in Korean could have similar meaning with more than 10 variations in the suffix as shown in Table \ref{agglutinative_example}.

Various language modeling methods that rely on character or morpheme like segmentation of words have been developed \cite{ciloglu2004language, cui2014learning, kim2016character, mikolov2012subword, zheng2013deep, ling2015finding}. \cite{chen2015joint} explored the idea of joint training for character and word embedding. Morpheme based segmentation has been explored in both Large Vocabulary Continuous Speech Recognition (LVCSR) tasks for Egyptian Arabic \cite{mousa2013morpheme} and German newspaper corpus \cite{cotterell2015morphological}.  \cite{sennrich2015neural} used subword units to perform machine translation for rare words. 

Morpheme distribution has a relatively smaller frequency tail as compared to the word distribution from vocabulary, hence avoids over-fitting for tail units. However, even with morpheme segmentation the percentage of out-of-vocabulary (OOV) words is significantly high in Korean. Character embedding in Korean is unfeasible as the context of the word is not sufficiently captured by the long sequence which composes the word. We select as features syllable-level embedding which has shorter sequence length and morpheme-level embedding to capture the semantics of the word. 

We deploy our model for input word prediction on mobile devices. To achieve desirable performance we are required to create a model that has as small as possible memory and CPU footprint without compromising its performance. We use differentiated softmax \cite{chen2015strategies} for the output layer. This method uses more parameters for the words that are frequent and less for the ones that occur rarely. We achieve better performance than existing approaches in terms of Key Stroke Savings (KSS) \cite{fowler2015effects} and our approach has been commercialized.

\begin{table}
\begin{center}
\begin{tabular}{ llc }
\hline
\hline
 Word & Morpheme & English \\ 
 \hline
 그가 & 그 + 가 & he \\  
 그는 & 그 + 는 & he \\
 그에게 & 그 + 에게 & to him \\
 그도 & 그 + 도 & him(he) also \\
 그를 & 그 + 를 & him \\
 그의  & 그 + 의 & his\\
 \hline
\end{tabular}
\end{center}
\caption{Example of variation of a base word  `그(He)'. It can have more than 10 variation forms according to its postposition.}
\label{agglutinative_example}
\end{table}

\section{Proposed Method}

\begin{figure}[t]
  \includegraphics[width=\columnwidth]{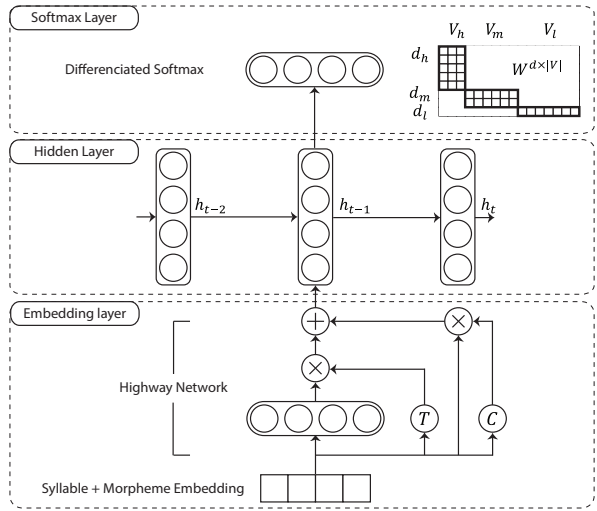}
  \centering
\caption{Overview of the proposed method. $T$ and $C$ are the transform gate and carry gate of the highway network respectively}
\label{overview_model}
\end{figure}

Following sections propose a model for agglutinative language. In Section \ref{basic-lm} we discuss the basic architecture of the model as detailed in Figure \ref{overview_model}, followed by Section \ref{syll-morph} that describes our embeddings. In Section \ref{diff-softmax} we propose an adaptation of differentiated softmax to reduce the number of model parameters and improve computation speed.

\subsection{Language Model} \label{basic-lm}

Overall architecture of our language model consists of a) embedding layer, b) hidden layer, c) softmax layer. Embedding comprises of syllable-level and morpheme-level embedding as described in Section \ref{syll-morph}. We combine both embedding features and pass them through a highway network \cite{srivastava2015highway} which act as an input to the hidden layers. We use a single layer of LSTM as hidden units with architecture similar to the non-regularized LSTM model by \cite{zaremba2014recurrent}. The hidden state of the LSTM unit is affine-transformed by the softmax function, which is a probability distribution over all the words in the output vocabulary.

\begin{figure}[t]
  \includegraphics[width=\columnwidth]{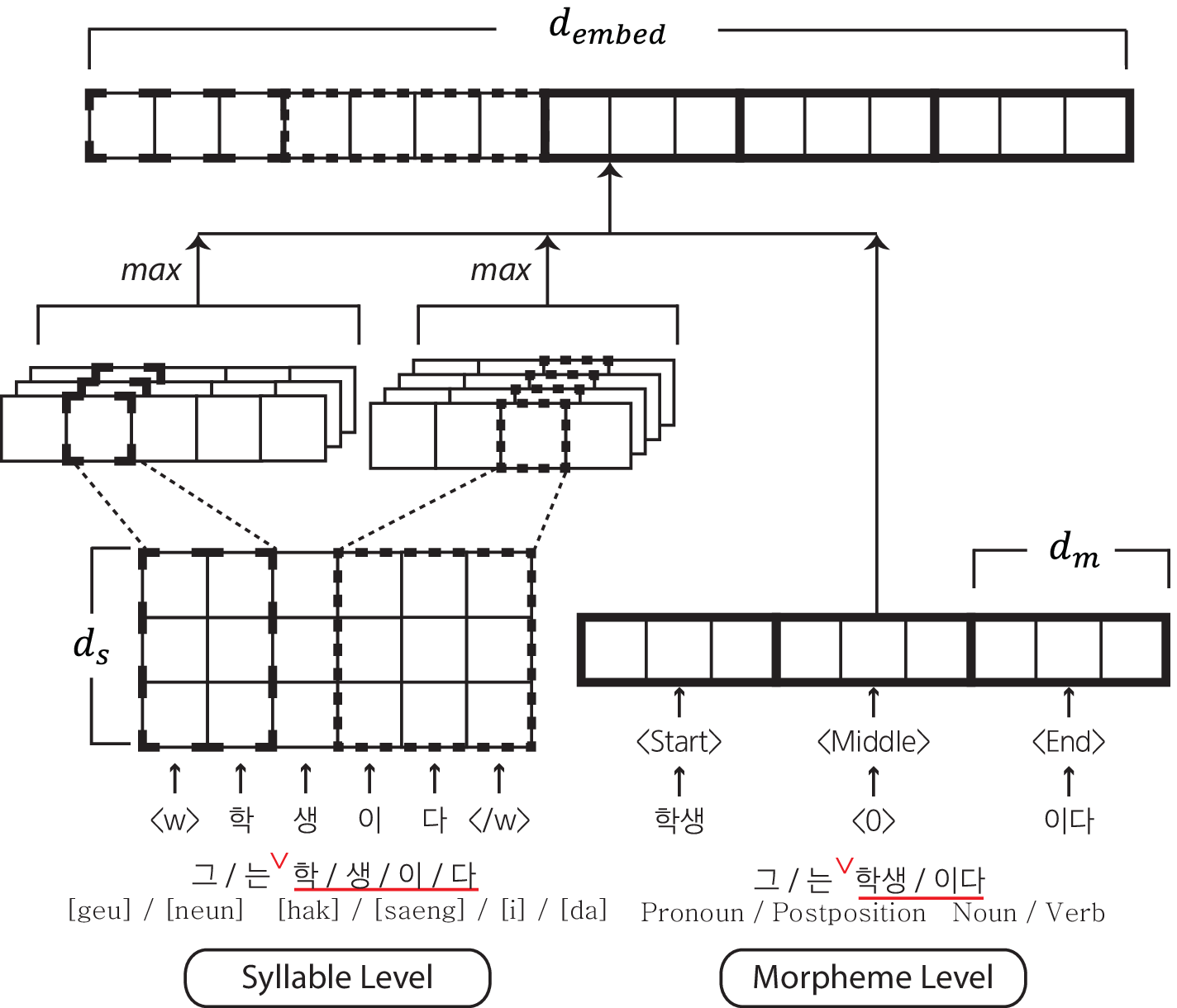}
  \centering
\caption{Proposed embedding method for agglutinative languages. We take an input word as syllable and morpheme level, embed them separately and concatenate them to make an entire embedding.}
\label{syllable_morpheme_embedding}
\end{figure}

\subsection{Syllable \& Morphological Embedding} \label{syll-morph}
We propose syllable-level embedding that attenuates OOV problem. \cite{santos2014learning, kim2016character} proposed character aware neural networks using convolution filters to create character embedding for words. We use convolution neural network (CNN) based embedding method to get syllable-level embedding for words. We use 150 filters that consider uni, bi, tri and quad syllable-grams to create a feature representation for the word. This is followed by max-pooling to concatenate the features from each class of filters resulting in a syllable embedding representation for the word. Figure \ref{syllable_morpheme_embedding} in the left half shows an example sentence embedded using the syllable-level embedding.

\begin{figure}[t]
  \includegraphics[width=\columnwidth]{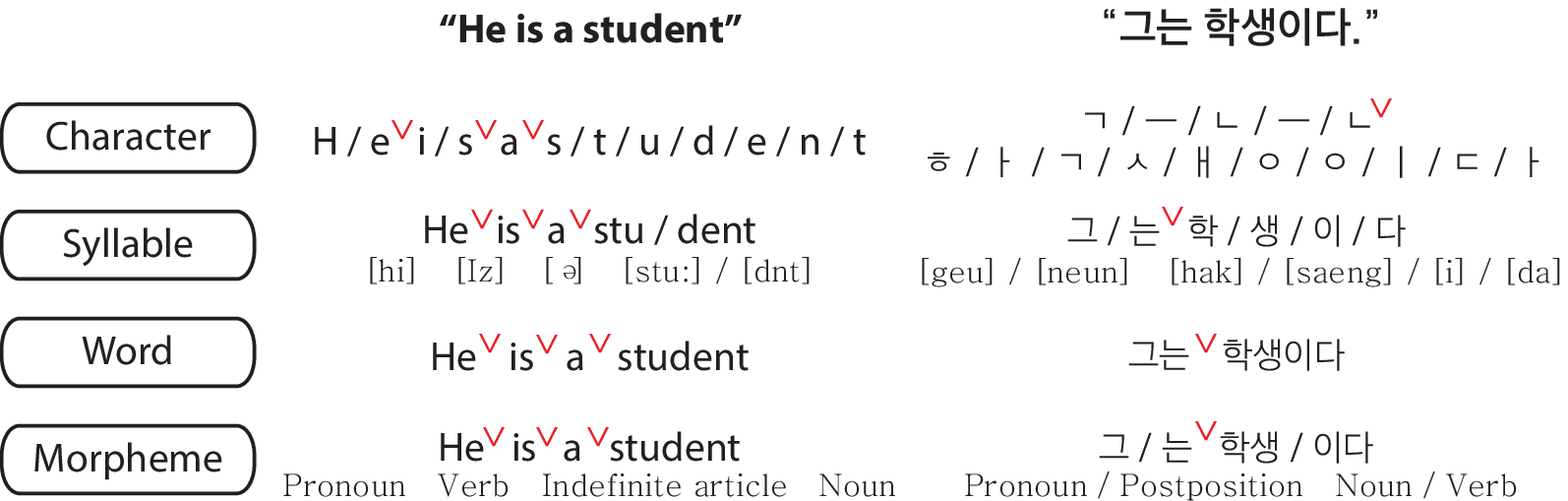}
  \centering
\caption{Comparison of various embedding levels. In case of Korean, syllable can be used as a basic unit of sequence to solve OOV with shorter sequence length compare to character level. Also, morpheme level is effective to make the size of vocabulary smaller.}
\label{embedding_examples}
\end{figure}

Figure \ref{embedding_examples} highlights the difference between various embedding and the features they capture. The syllable embedding is used along with a morphological embedding to provide richer features for the word. The majority of words (95\%) in Korean has at most three morphological units. Each word can be broken into start, middle, and end unit. We embed each morphological unit by concatenating to create a joint embedding for the word. Advantage of morphological embedding over syllable is all the sub-units have an abstract value in the language and this creates representation for words relying on the usage of these morphemes. Both morphological and syllable embeddings are concatenated and fed through a highway network \cite{srivastava2015highway} to get a refined representation for the word as shown in the embedding layer for Figure \ref{overview_model}.

\subsection{Differentiated Softmax} \label{diff-softmax}

The output layer models a probability distribution over words in vocabulary conditioned on the given context. There is a trade-off between required memory and computational cost which determines the level of prediction. To generate a complete word, using morpheme-level predictions requires beam search which is expensive as compared to word-level predictions. Using beam search to predict the word greedily does not adhere to the computational requirements set forth for mobile devices. Thus, we have to choose word-level outputs although it requires having a vocabulary of over 0.2M words to cover 95\% of the functional word forms. Computing a probability distribution function for 0.2M classes is computational intensive and overshoots the required run-time and the allocated memory to store the model parameters. 

Therefore, the softmax weight matrix, $W_{softmax}$, needs to be compressed as it is contributing to huge model parameters. We initially propose to choose an appropriate rank for the $W_{softmax}$ in the following approximation problem; $W_{softmax} = W_{A} \times W_{B}$, where $W_{A}$ and $W_{B}$ have ranks less than $r$. We extend the idea of low rank matrix factorization in \cite{sainath2013low} by further clustering words into groups and allowing a different low rank $r'$ for each cluster. The words with high frequency are given a rank, $r_{1}$, such that $r_{1} \ge r_{2}$ where $r_{2}$ is the low rank for the words with low frequency. The core idea being, words with higher frequency have much richer representation in higher dimensional space, whereas words with low frequency cannot utilize the higher dimensional space well. 

We observe that 87\% of the words appear in the tail of the distribution by the frequency of occurrence. We provide a higher rank to the top 2.5\% words and much lower rank to the bottom 87\%. This different treatment reduces the number of parameters and leads to better modeling.


\section{Experiment Results}
\subsection{Setup}
We apply our method to web crawled dataset consisting on news, blogs, QA. Our dataset consists of over 100M words and over 10M sentences. For morpheme-level segmentation, we use lexical analyzer and for syallable-level we just syllabify the dataset. We empirically test our model and its input vocabulary size is around 20K morphemes and 3K syllables. The embedding size for morpheme is 52 and that for syllable is 15. We use one highway layer to combine the embeddings from syllable and morpheme. Our hidden layer consists of 500 LSTM units. The differentiated softmax outputs the model's distribution over the 0.2M words in the output vocabulary with top 5K (by frequency) getting a representation dimension (low rank in $W_{softmax}$) of 152, next 20K use a representation dimension of 52 and the rest 175K get a representation dimension of 12. All the compared models have word level outputs and use differentiated softmax.

\subsection{Comparison of embedding methods}

We randomly select 10\% of our crawled data (10M words, 1M sentences) to compare embedding methods as shown in Table \ref{perplexity_results}. We test character, syllable, morpheme and word-level embeddings. The word-level embedding has the highest number of parameters but has the worst performance. As expected breaking words into their sub-forms improves the language model. However, our experiment reaches its peak performance when we use syllable level embeddings. To improve the performance even further we propose using syllable and morpheme which outperforms all the other approaches in terms of perplexity.

\begin{table}
\begin{center}
\begin{tabular}{ lrrr }
\hline
\hline
 Embedding & Param. & Perplexity & Vocab.\\ 
 \hline
 Word & 15.72M & 327.17 & 200K \\
 Morph & 6.61M & 283.54 & 20K\\
 Character & 8.66M & 235.52 & 40\\  
 \hline
 Syl & 8.71M & 231.30 & 3K\\
  \textbf{Syl + Morph} & \textbf{9.50M} & \textbf{218.65} & \textbf{23K} \\
 \hline
\end{tabular}
\end{center}
\caption{Results of different embedding methods. Param. : Total model paramerters, Vocab: Input vocabulary size, Syl : Syllable, Morph: Morpheme.}
\label{perplexity_results}
\end{table}

\subsection{Performance evaluation}
Proposed method shows the best performance compared to other solutions in terms of Key Stroke Savings (KSS) as shown in Table \ref{KSS_experiments}. KSS is a percentage of key strokes not pressed compared to a vanilla keyboard which does not have any prediction or completion capabilities. Every user typed characters using the predictions of the language model counts as key stroke saving. The dataset\footnote{The dataset consists of 67 sentences (825 words, 7,531 characters) which are collection of formal and informal utterances from various sources. It is available at \url{https://github.com/meinwerk/SyllableLevelLanguageModel}} used to evaluate KSS was manually curated to mimic user keyboard usage patterns.

The results in Table \ref{KSS_experiments} for other commercialized solutions are manually evaluated due to lack of access to their language model. We use three evaluators from inspection group to cross-validate the results and remove human errors. Each evaluator performed the test independently for all the other solutions to reach a consensus. We try to minimize user personalization in predictions by creating a new user profile while evaluating KSS.  

\begin{figure}[t]
  \includegraphics[width=\columnwidth]{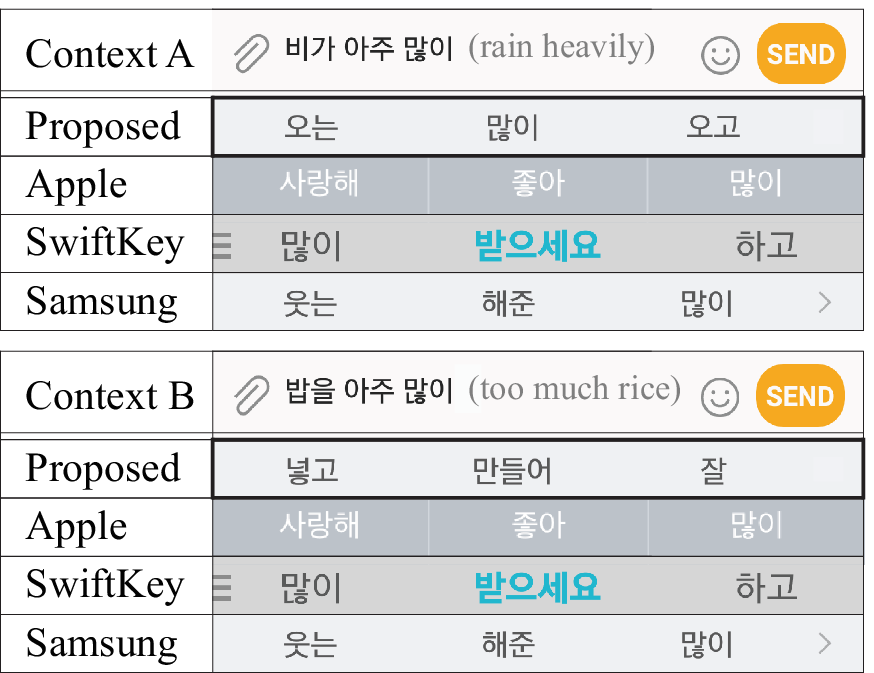}
  \caption{Example of comparison with other commercialized solutions. Predicted words for the Context A (rain heavily) and Context B (too much rice). Other solutions make same prediction regardless of the context (only consider the last two words of context).}
\label{keyboard}
\end{figure}

The proposed method shows 37.62\% in terms of KSS and outperforms compared solutions. We have achieved more than 13\% improvement over the best score among existing solutions which is 33.20\% in KSS. If the user inputs a word with our solution, we require on an average 62.38\% of the word prefix to recommend the intended word, while other solutions need 66.80\% of the same. Figure \ref{keyboard} shows an example of word prediction across different solutions. In this example, the predictions from other solutions are same irrespective of the context, while the proposed method treats them differently with appropriate predictions.

\begin{table}[t]
\begin{center}
\begin{tabular}{ lcc }
\hline
\hline
Developer & KSS(\%) \\
\hline
Proposed  & {\bf 37.62} \\
Swiftkey & 33.20 \\
Apple  & 31.90 \\
Samsung  & 31.40 \\
 \hline
\end{tabular}
\end{center}
\caption{Performance comparison of proposed method and other commercialized keyboard solutions by various developers.}
\label{KSS_experiments}
\end{table}

\section{Conclusion}	
We have proposed a practical method for modeling agglutinative languages, in this case Korean. We use syllable and morpheme embeddings to tackle large portion of OOV problem owing to practical limit of vocabulary size and word-level prediction with differentiated softmax to compress size of model to a form factor making it amenable to running smoothly on mobile device. Our model has 9.50M parameters and achieves better perplexity than character-level embedding by 16.87. Our proposed method outperforms the existing commercialized keyboards in terms of key stroke savings and has been commercialized. Our commercialized solution combines above model with n-gram statistics to model user behavior thus supporting personalization.

\bibliography{emnlp2017}
\bibliographystyle{emnlp_natbib}

\end{document}